\documentclass[11pt]{article}

\usepackage[margin=1in]{geometry}
\usepackage{booktabs}
\usepackage{multirow}
\usepackage{comment}
\usepackage{caption}
\usepackage{amsmath, amssymb}
\usepackage{tikz}
\usetikzlibrary{arrows.meta, positioning, calc, fit}
\usepackage{graphicx}
\usepackage[hidelinks]{hyperref}

\usepackage{authblk}

\usepackage[authoryear]{natbib}
\newcommand{\orcid}[1]{\href{https://orcid.org/#1}{ORCID: #1}}

\begin{document}
\thispagestyle{empty}


\title{From ML Predictions to Informed Diagnostic Assistance using the Toulmin Model of Argumentation}

\author[1]{Anca Marginean \orcid{0000-0001-8426-588X}}

\author[1]{Adrian Groza \orcid{0000-0003-0143-5631}}

\affil[1]{Artificial Intelligence Research Institute AIRi@UTCN, \\
Technical University of Cluj-Napoca, Romania}

\date{} 
\maketitle
\thispagestyle{empty} 

\begin{abstract}
To provide a structured and interpretable assessment, we decompose the image-based diagnosis into components following the Toulmin model of argumentation. This model consists of a claim, grounds, warrant, qualifier, rebuttal, and backing.
Consider a claim generated by a machine learning (ML) model for retinal diagnosis. Rather than accepting this claim at face value, one could either apply explainable AI (XAI) methods or adopt an argumentation-based approach.
In our framework, a model specialized in biomarker extraction from images provides the grounds. The warrant—linking the grounds to the claim—is analyzed by an agent equipped with medical knowledge; in our architecture, this role is fulfilled by a MedGemma agent.
The qualifier is determined based on the overall quantitative evaluation of both the warrant and grounds models. Finally, a rebuttal is constructed using image similarity measures computed with MedSigLip.
All these components are presented to the human expert, enabling a more informed and critical assessment of the ML-generated diagnosis.
\end{abstract}

\noindent\textbf{Keywords:} Toulmin model of argumentation; MedGemma; retinal diagnosis; object detection 



\section{Introduction}
There is increasing interest in AI approaches that provide transparent and evidence-based reasoning. Argumentation frameworks are well-suited to this objective because they explicitly connect observations, assumptions, and conclusions while also allowing uncertainty and counter-evidence to be represented. The Toulmin model of argumentation is particularly relevant, as it structures reasoning through components such as claim, grounds, warrant, qualifier, rebuttal, and backing.

In this work, we explore the use of the Toulmin model for a multimodal retinal image assessment system. Our approach combines automated biomarker extraction from OCT scans, retrieval of visually similar reference cases, and reasoning performed by a medical large language model. By organizing these heterogeneous sources of information into a coherent argumentative structure, the system aims to support clinicians with a more interpretable assessment process.
We evaluated the proposed system on two tasks: T1) given an OCT image, determine whether the patient suffers from age-related macular degeneration (AMD); T2) given an OCT image from a patient with AMD, identify the stage of the disease.

\section{Reference cases}
In order to build our set of reference cases, we used cases from the OCT Training Manual provided by Heidelberg, the manufacturer of the OCT acquisition device.

\begin{figure}[t]
    \centering
    \includegraphics[width=0.25\linewidth]{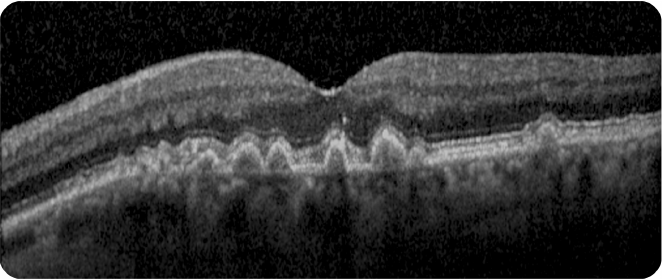}
    \hspace{2cm}
    \includegraphics[width=0.25\linewidth]{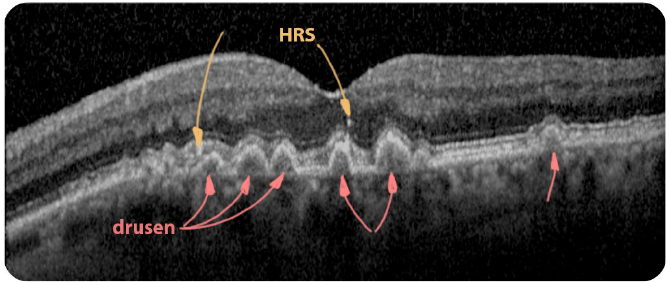}
    \includegraphics[width=0.45\linewidth]{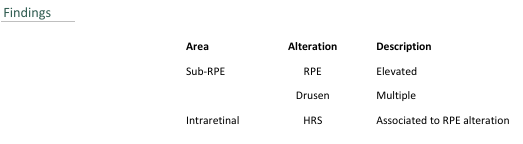}
    \hspace{1cm}
    \includegraphics[width=0.45\linewidth]{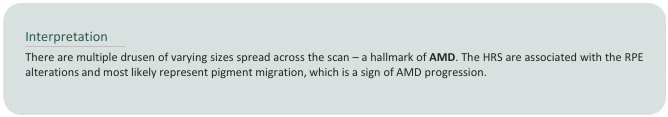}
    \caption{Case 3 included in the OCTTraining document: the raw OCT image, the structured $Findings$, visualisation of the alterations and $Interpretation$ }
    \label{fig:octtraining}
\end{figure}

Each case includes a fundus image, an OCT image, a list of findings corresponding to a structured OCT assessment, a clinical interpretation, and a visually annotated image highlighting the detected alterations.
These alterations may be located in different areas of retina: sub-RPE, subretinal, intraretinal, and epi-/preretinal.
For each identified alteration, a textual description is provided. Figure \ref{fig:octtraining} presents an example of such a case (Case 3). 

We extracted all the cases and we built the \emph{RefC$\_$dataset} with $93$ cases. 
We selected the OCT training manual because it serves as a reference guide for OCT assessment and is not focused on any single retinal condition. In addition, unlike most publicly available OCT datasets, these cases provide detailed information on the appearance of the alterations together with clear visual annotations indicating their localization and an interpretation. Together, this information can be highly informative for the human expert.
The purpose of this dataset is to support the diagnostic process in two respects: first, by helping to validate or refine biomarker identification, and second, by supporting the reasoning that links these biomarkers to a diagnosis or staging information. 

\section{Biomarker extraction models}
\begin{table}[t]
\footnotesize
    \centering
    \caption{Subset of evaluations for YOLOE models extracted from \cite{Ardelean2025}}
    \begin{tabular}{lccc||lccc}
      
    \toprule
    \multicolumn{4}{c||}{YOLOE trained on AROI} & \multicolumn{4}{c}{YOLOE trained on OCT5k}\\
    \midrule
          Class &Precision & Recall & mAP@50 & Class &Precision & Recall & mAP@50\\
          \midrule
             PED & 0.76 & 0.62 & 0.74   & Choroidalfolds & 0.43 & 0.3 & 0.40\\
         SRF & 0.86 & 0.74 & 0.84   & SoftDrusen     & 0.74 & 0.38 & 0.56\\
         IRF & 0.73 & 0.44 & 0.59   & HardDrusen     & 0.43 & 0.25 & 0.34 \\
         &&&&                         Geographic Atrophy & 0.83 & 0.56 & 0.72\\
         \bottomrule
    \end{tabular}
  
    \label{tab:yoloe}
\end{table}

We use two publicly available models for biomarker extraction\cite{Ardelean2025}. Both models are based on YOLOE and were trained on two different OCT datasets: AROI \cite{aroi} and OCT5K \cite{oct5k}. 

The AROI dataset contains OCT scans from patients with AMD, whereas OCT5k covers a broader range of retinal pathologies, including AMD and DME (diabetic macular edema). 
The YOLOE trained on AROI detects three biomarkers: SRF(subretinal fluid), IRF(intraretinal fluid), and PED (pigment epithelial detachment). 
The model trained on OCT5k targets nine biomarkers, including soft drusen, hard drusen, retinal fluid, geographic atrophy, and choroidal folds. 
According to the evaluations reported in \cite{Ardelean2025} (Table \ref{tab:yoloe}), YOLOE trained on AROI achieves good performances for PED, SRF and IRF detection, although very small IRF regions may occasionally be missed.
In contrast, the model trained on OCT5k shows lower overall performance.

\begin{figure}
    \centering
    \includegraphics[width=0.55\linewidth]{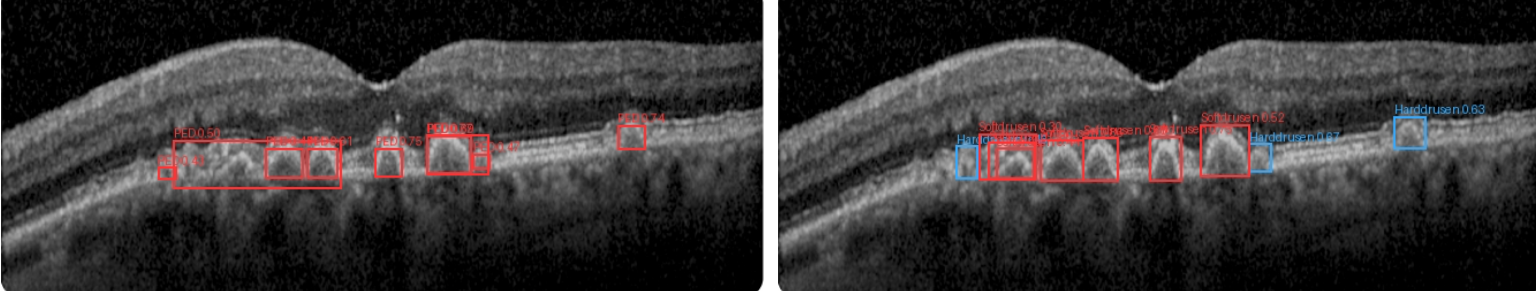}
    \caption{YOLOE models (left: AROI, right: OCT5k) detections for the Case 3 (Figure \ref{fig:octtraining}). For AROI model, Red = PED. For the OCT5k model, Blue = Hard Drusen, and Red = SoftDrusen.}
    \label{fig:yoloepreds}
\end{figure}

An important observation is the relation between PED and drusen: once the size of a drusen increases drastically, or multiple drusens merge, the resulting alteration is called drusenoid PED. There are two other types of PED: fibrovascular (F-PED) and serous PED (S-PED). The AROI dataset, consequently, the AROI model, does not differentiate between PED types and tends to consider drusens of any size as PED. This observation is very important since the staging of AMD depends on the presence or absence of PED. Figure \ref{fig:yoloepreds} shows the detections of YOLOE AROI and OCT5k model on the case included in Figure \ref{fig:octtraining}. It can be observed that the OCT5k model correctly identifies drusens (hard or soft), while the AROI model considers all the drusens to be PED. 

\begin{figure}
\begin{minipage}[t]{0.46\textwidth}
\vspace{0pt}
\footnotesize
    \centering
    \captionof{table}{Image level evaluation of YOLOE AROI model on reference cases \emph{RefC$\_$Dataset}}
    \begin{tabular}{lll}
    \toprule
    Class & Precision (0.25 $\rightarrow$ 0.5) &  Recall (0.25 $\rightarrow$ 0.5)\\
    \midrule
         PED& 0.533 $\rightarrow$ 0.548 & 0.923 $\rightarrow$ 0.654 \\
         IRF& 0.923 $\rightarrow$ 1 & 0.857 $\rightarrow$ 0.714 \\
         SRF& 0.677 $\rightarrow$ 0.762 & 0.778 $\rightarrow$ 0.593 \\
         \bottomrule
    \end{tabular}
    
    \label{tab:prec_ref}
\end{minipage}
\hfill
\begin{minipage}[t]{0.52\textwidth}
\vspace{0pt}
\centering
\includegraphics[width = 0.48\linewidth]{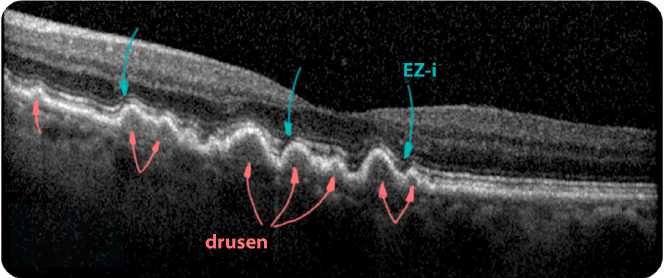} \hfill
\includegraphics[width=0.48\linewidth]{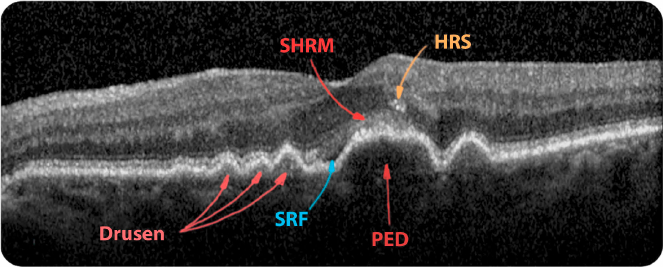}
\captionof{figure}{The closest cases to Case 3: Case 28 (left) $.955$ similarity. Case 13 (right) $.954$ similarity. }
\label{fig:case28}
\end{minipage}

\end{figure}

We evaluated YOLOE AROI model on the reference cases. In our reference cases, no annotated bounding boxes are available. Therefore, we consider PED, IRF, SRF to be present whenever the corresponding biomarker is mentioned in given findings. Image-level precision and recall are reported in Table \ref{tab:prec_ref}. As expected, PED shows low precision but high recall. The model frequently confuses drusen or other RPE changes with PED, which is anatomically plausible since both involve RPE elevation, and the AROI dataset does not explicitly emphasize the distinction between drusen and PED. We also observe that the number of PED false negatives at a threshold of $0.5$ increases substantially compared to a threshold of $0.25$. A manual review of these FN samples showed that PED was consistently present together with other alterations located near the RPE. These observations are later incorporated into the construction of our qualifier.

\section{System architecture}
\begin{figure}
    \centering
    \includegraphics[width=0.92\linewidth]{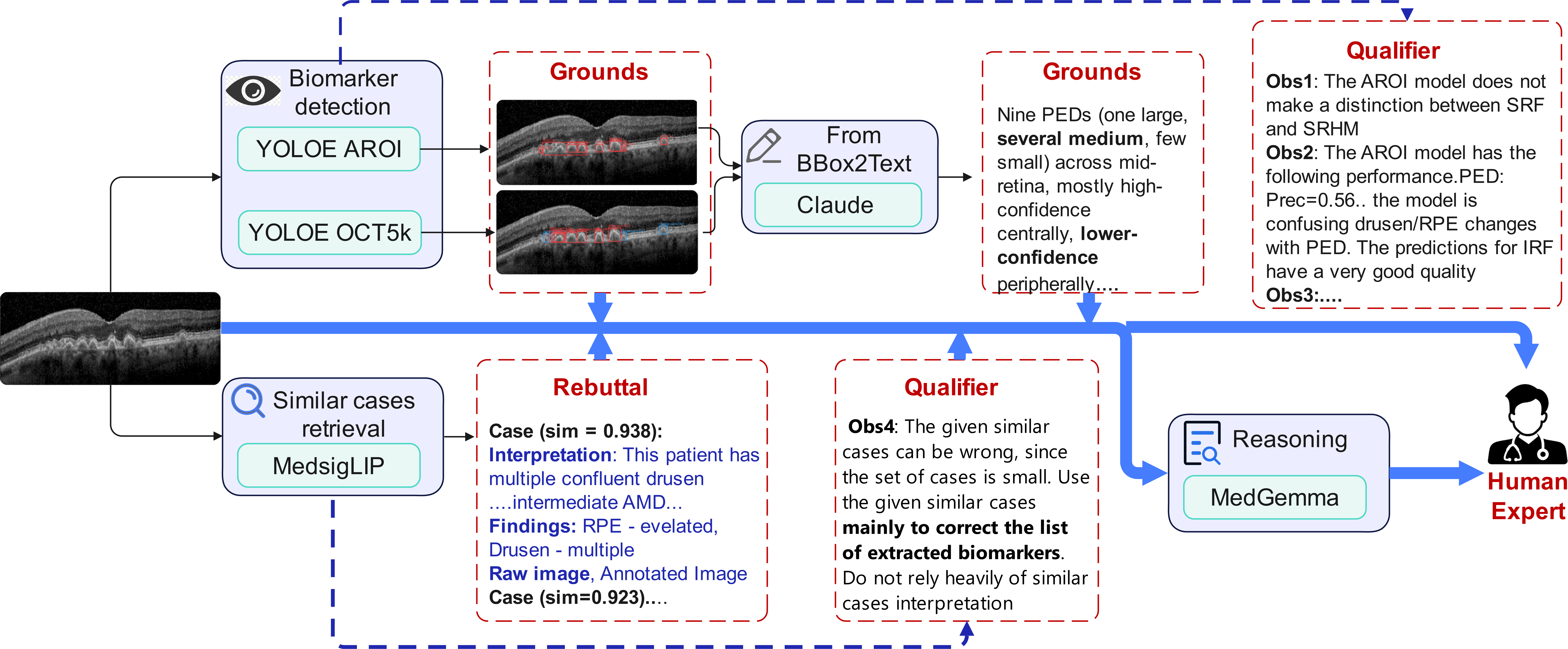}
    \caption{System architecture}
    \label{fig:system}
\end{figure}

We propose a system (Figure \ref{fig:system}) that 
(i) detects evidence in the input image and summarizes the detections (grounds), 
(ii) retrieves visually similar cases from a reference case dataset (rebuttal), and 
(iii) revives a textual qualifier based on the overall evaluation of the detection and retrieval modules. 
This information is then provided to MedGemma for reasoning. Both the extracted evidence and the generated reasoning are presented to the human expert.

MedGemma is a medically adapted version of the Gemma LLM. 
The multimodal variants include a SigLIP image encoder pretrained on medical data, including ophthalmology images (mostly eye fundus). The LLM component was trained not only on medical question-answer pairs but also on medical text, health records, and medical record comprehension tasks. We use the 4-bit quantization of the multimodal variant 27B from Huggingface, without any additional finetuning. 

\subsection{Grounds - Biomarker detection models}
The grounds for our reasoning are extracted with two YOLOE detection models. 
Raw YOLOE detections (BBoxes, confidence, class, and computed area) are translated into concise, physician-friendly language using Claude Opus 4.7, based on number, relative position, confidence interpretation.
For example, the AROI model detections from Fig. \ref{fig:yoloepreds} are summarized as \emph{Nine PEDs (one large, several medium, few small) across mid-retina, mostly high-confidence centrally, lower-confidence peripherally.} 
The MedGemma provided text include both the raw detections and the summarized ones.

\subsection{Rebutals - Similarity Case Retrieval (MedSigLIP)}
MedSigLIP is a medical-domain adaptation of the SigLIP vision–language model designed to align medical images and text within a shared embedding space. It is not particularly trained on OCT images. 
We use MedSigLIP to retrieve OCT reference cases that are visually similar to a query scan.
For each reference case, the MedSigLIP image embeddings, the $Findings$, and the $Interpretation$ are all stored in OpenSearch. The advantage of using OpenSearch is that three retrieval models are possible: vector-only, text-only, and hybrid.

For example, when we use vector-only retrieval for Case 3 (Fig. \ref{fig:octtraining}) from our \emph{RefC$\_$dataset}, the closest 2 images correspond to Case 28 and Case 13 (Fig. \ref{fig:case28}). 
We can observe that the Findings are similar. However, when analyzing the Interpretations, it becomes clear that small differences in the images correspond to significant differences in interpretation.
All three cases have AMD as a diagnosis, but the stage of the disease is probably \emph{Late} for Case 3 and Case 13, while for Case 28 it is \emph{Intermediate}. 

\subsection{Qualifier - Overall performance evaluations}
As a qualifier component of our argument, we consider the following observations.
$O_1$: \emph{The AROI model does not make a distinction between SRF and SRHM (subretinal hyperreflective material), consequently they were jointly annotated as SRF.}
$O_2$: \emph{The AROI model has the following performance}...(the values from Table \ref{tab:prec_ref}).
\emph{Consequently, the model confuses the drusen/RPE changes with PED. The predictions for IRF are of very good quality. }
$O_3$: \emph{The OCT5k model has very good precision on Softdrusen and SoftdrusenPED but a smaller recall.}
$O_4$: \emph{The given similar cases can be wrong, since the set of cases is small. Use the given similar cases mainly to correct the list of extracted biomarkers; do not rely heavily on the interpretation of similar cases.}


\section{Experiments and Evaluation}

For the automated quantitative evaluation of our framework, five experiments were performed on $200$ cases extracted from the OCTDL dataset \cite{Kulyabin2024}. We used $100$ images from the AMD class, and $100$ from the DME class. For the AMD diagnosis, OCTDL also includes annotation on staging: early, intermediate, and late.

The first experiment ($Exp_0$) measured the ability of MedGemma to solve T1 and T2 without additional information besides the OCT scan. 
The next four experiments differ in terms of rebuttal and qualifier. 
In $Exp_1$, the rebuttal includes the top-3 cases retrieved but only if the similarity is above $0.93$. The qualifier is not included.
In $Exp_2$, the rebuttal is similar to $Exp_1$, but we include the first three observations for the qualifier.
In $Exp_3$, the qualifier is the same as in $Exp_2$, but no similar cases are provided (therefore no rebuttal).
In $Exp_4$, the rebuttal includes the top-3 cases without the threshold-based condition. The qualifier includes all four observations.

Table \ref{tab:results} reports the precision and recall for T1 and T2. It can be observed that MedGemma alone  achieves the weakest overall performance on both T0 and T1, indicating that providing externally extracted information is necessary. The contribution of similar cases retrieved suggests room for improvement.
For T1, $Exp_1$ and $Exp_3$ - where similar cases are restricted or absent- achieve the best balance between precision and recall. 
For T2, however, rebuttals combined with a cautious qualifier appear to provide benefit particularly for late-stage AMD.
Early AMD obtains the weakest performance due to the fact that AROI model does not differentiate between drusens and PED, and in AMD, PED is a significant finding that moves the diagnosis beyond early AMD.
\begin{table}[t]
\footnotesize
\centering
\caption{Quantitative evaluations of the system on T1 and T2}
\label{tab:results}

\begin{tabular}{lccc|ccc}
\toprule
\multirow{2}{*}{Experiment} & \multicolumn{3}{c|}{T1 - Is it AMD?} & \multicolumn{3}{c}{T2 - AMD staging} \\
& Precision & Recall & FP & Early P/R & Intermediate P/R & Late P/R \\
\midrule
$Exp_0$ & 0.56  & 0.39  & 59 & 0.12/0.00  & 0.27 / 0.2 & 0.55/ 0.66 \\
$Exp_1$ & 0.79  & 0.93  & 24 & 0 / 0      & 0.27 / 0.3 & 0.64 / 0.82 \\
$Exp_2$ & 0.54  & 0.97  & 82 & 0 / 0 & 0.21 / 0.30 & 0.66 / 0.79 \\
$Exp_3$ & 0.76  & 0.90  & 27 & 0.20 / 0.05 & 0.17 / 0.25 & 0.69 / 0.776 \\
$Exp_4$ & 0.604 & 0.957 & 59 & 0 / 0 & 0.26 / 0.31 & 0.725 / 0.943 \\
\bottomrule
\end{tabular}
\end{table}

\section{Discussion and Conclusions}
The notions of argumentation, justification, and explanation are related, but not the same. 
Argumentation is the broader process of defending a position in the presence of possible disagreement.
Justification is narrower: it provides support for accepting a position, while explanation increases understanding of a position, process, or event.

The complementarity between argumentation and explanation is characterised by the fact that clinicians base their decisions on both evidence and understanding \cite{letia2012interleaved}. For example, image-derived biomarkers provide evidence supporting a diagnostic claim, but this evidence is more meaningful when accompanied by an explanation of how these findings relate to the underlying retinal pathology. Conversely, an explanation of the disease mechanism alone is insufficient without plausible visual evidence.
Zhan et al. \cite{zhan2026answersargumentstrustworthyclinical} propose a training pipeline in which Curriculum Goal-Conditioned
Learning (CGCL) trains LLMs to generate diagnostic arguments
that explicitly follow this Toulmin structure. 
However, it focuses primarily on text-based clinical reasoning and LLM training.

In contrast, we consider that not only the output reasoning, but also the information provided to the LLM for a given case should be organized according to the Toulmin structure. By adopting such a framework, external models can be integrated as complementary sources of evidence, extending the LLM’s internal knowledge with image-based findings and other specialized analyses.
In our case, the multimodal LLM (MedGemma) was not specifically trained on OCT scans, but it has knowledge of the anatomy of the retina, the normal structure of the retina, and pathological alterations.

Our work is closely aligned with the growing view that AI in healthcare should expose reasoning processes, not only final answers.
 YOLOE models extract evidence (grounds) in the form of detectable biomarkers that can be directly inspected. Retrieved similar cases (rebuttal) may support or challenge these findings. The qualifier is derived from the overall performance and confidence of the detection and retrieval modules. As a result, MedGemma receives a structured and interpretable input that can also be reviewed by the human expert.

The proposed framework is general and not limited to diagnosis based on OCT imaging. The main limitations of our current work are the small number of reference cases, the limited evaluation, and the absence of subjective assessment by a human expert.

\small{
\textbf{Acknowledgment.} This work was supported in part by the project ”Romanian Hub for Artificial Intelligence-HRIA” Smart Growth, Digitization and Financial Instruments Program, MySMIS no. 334906.
}
\bibliography{references}

\end{document}